# Deep learning incorporating biologically-inspired neural dynamics

Stanisław Woźniak[1,*], Angeliki Pantazi[1], Thomas Bohnstingl[1,2], and Evangelos Eleftheriou[1]

[1] IBM Research – Zurich, Rüschlikon, Switzerland
[2] Institute of Theoretical Computer Science, Graz University of Technology, Graz, Austria

**Neural networks have become the key technology of artificial intelligence and have contributed to breakthroughs in several machine learning tasks, primarily owing to advances in deep learning applied to Artificial Neural Networks (ANNs). Simultaneously, Spiking Neural Networks (SNNs) incorporating biologically-feasible spiking neurons have held great promise because of their rich temporal dynamics and high-power efficiency. However, the developments in SNNs were proceeding separately from those in ANNs, effectively limiting the adoption of deep learning research insights. Here we show an alternative perspective on the spiking neuron that casts it as a particular ANN construct called Spiking Neural Unit (SNU), and a soft SNU (sSNU) variant that generalizes its dynamics to a novel recurrent ANN unit. SNUs bridge the biologically-inspired SNNs with ANNs and provide a methodology for seamless inclusion of spiking neurons in deep learning architectures. Furthermore, SNU enables highly-efficient in-memory acceleration of SNNs trained with backpropagation through time, implemented with the hardware in-the-loop. We apply SNUs to tasks ranging from hand-written digit recognition, language modelling, to music prediction. We obtain accuracy comparable to, or better than, that of state-of-the-art ANNs, and we experimentally verify the efficacy of the in-memory-based SNN realization for the music-prediction task using 52,800 phase-change memory devices. The new generation of neural units introduced in this paper incorporate biologically-inspired neural dynamics in deep learning. In addition, they provide a systematic methodology for training neuromorphic computing hardware. Thus, they open a new avenue for a widespread adoption of SNNs in practical applications.**

Research interest in neural networks has considerably grown over recent years owing to their remarkable success in many applications. Record accuracy was obtained in deep learning of neural networks applied for image classification[1–3], multiple object detection trained end-to-end[4], pixel-level segmentation of images[5], language translation[6], speech recognition[7] and even in the playing of computer games based on raw screen pixels[8]. Although the term neural networks elicits associations to the sophisticated functioning of the brain, the advances in the field were obtained by extending the original simple ANN paradigm of the 50's to complex deep neural networks trained with backpropagation[9]. The ANNs take only high-level inspiration from the structure of the brain, comprising neurons interconnected with synapses, where their neural dynamics represent an input-output transformation through a non-linear activation function. ANNs are primarily used in applications that involve static data. However, to enable operation with sequential or temporal data, their dynamics have been extended to include recurrency, leading to so-called Recurrent Neural Networks (RNNs), and their powerful variants such as Long-Short-Term Memory (LSTM) units[10] and Gated Recurrent Units (GRUs)[11]. All these recent advances have resulted in human-like performance of ANNs for certain tasks, albeit at a much higher power budget than the ≈ 20 W required by the human brain.

At the same time, the neuroscientific community — whose focus is understanding the neural networks in the brain — has been exploring architectures with more biologically-realistic dynamics. The use of sparse asynchronous voltage pulses, called spikes, to compute and propagate information, in conjuction with the concept of spiking Leaky Integrate-and-Fire (LIF) neurons[12–14] led to the SNN paradigm. It was presented to solve interesting cognitive problems[15,16], model the competitive dynamics of inhibitory neural circuits[17], or exploit rich recurrent temporal dynamics[18]. Biologically-inspired learning rules, such as Spike-Timing Dependent Plasticity[19–21], were applied for correlation detection[20–26], high frequency signals sampling[22], handwritten digit recognition[27–29], or feature learning[30–32]. From an implementation perspective, the inherent characteristics of SNNs have led to highly-efficient computing architectures explored in the field of neuromorphic computing[33]. For example, incorporation of the sparse asynchronous temporal spiking neural dynamics and the physical collocation of neural processing and synaptic memory have led to the development of non-von Neumann systems with significantly increased parallelism and reduced energy consumption, demonstrated in chips such as FACETS/BrainScales[34], Neurogrid[35], IBM's TrueNorth[36], and Intel's Loihi[37]. Recent breakthroughs in the area of memristive nanoscale devices have enabled further improvements in area and energy efficiency of mixed digital-analog implementations of synapses and spiking neurons[22–25,38].

However, although SNNs are successful in few specific applications, they lack a general universal approach for quickly designing and training architectures suitable for many other applications. This made it unclear how to effectively scale them up to reach high accuracies for common machine learning tasks and to materialize the benefits of the low power neuromorphic hardware. Recent SNN advancements have focused on taking advantage of the scalable ANN training with backpropagation (BP) in two ways. Firstly, by porting the weights trained in ANNs[39–42] based on the similarity[12] of an averaged activity of an SNN to an ANN, but neglecting the timing of individual spikes that is crucial for low-power, low-latency handling of temporal problems. Secondly, by developing SNN learning approaches inspired by the idea of BP[43,44], exploring how STDP could perform BP[45,46], or how to directly derive BP for SNNs[47–51]. Despite many performance improvements, these approaches involve the additional effort of deriving and reimplementing deep learning insights specifically for SNNs.

* Corresponding author: stw@zurich.ibm.com

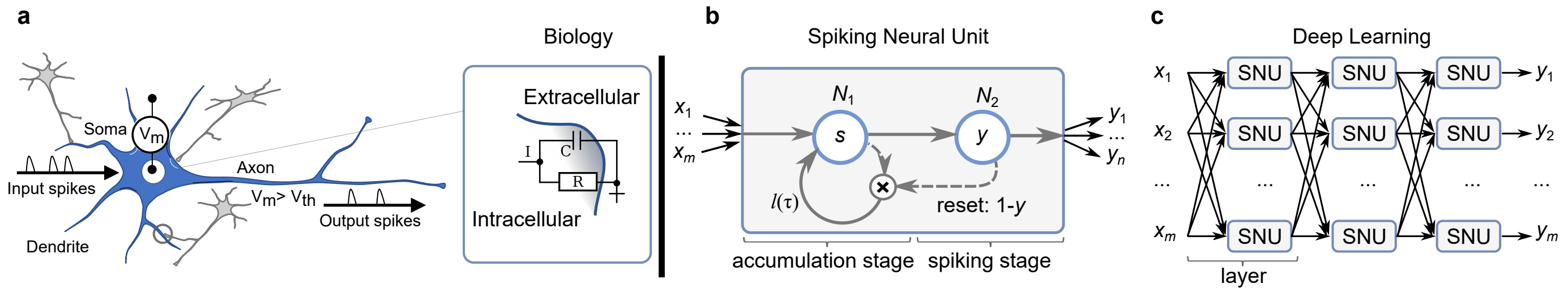

**Figure 1 | Spiking neural dynamics.** **a**, Biological neurons receive input spikes, which are integrated into the membrane potential $V_m$ of the soma and lead to emission of output spikes through the axon when $V_m$ crosses the spiking threshold $V_{th}$. This dynamics is often modelled using RC circuits. **b**, SNU models the spiking neural dynamics in the form of two ANN neurons, performing the integration and emission of output spikes. The neuron $N_2$ with a step activation function reproduces the spiking behaviour, whereas the neuron $N_2$ with a sigmoid activation function generalizes the neural dynamics beyond the spiking case. **c**, SNUs are common ANN units, so that they can be seamlessly applied to construct deep learning architectures.

Here, we take a different perspective of the relationship between ANNs and SNNs. We reflect on the nature of the temporal spiking neural dynamics and propose an approach that unifies the SNN model with its ANN counterpart. Firstly, we provide a constructive proof that a spiking neuron of LIF type can be transformed into a simple novel recurrent ANN unit called Spiking Neural Unit (SNU) and we further generalize it to a non-spiking case in a so-called soft variant of SNU (sSNU). Secondly, we discuss the consequences of this realization. Considering the ANN-SNN duality, we show that it is possible to reuse the existing ANN frameworks for seamless training of any SNN architecture with backpropagation through time. We demonstrate the efficacy of our approach by training deep SNNs with up to seven layers and analyzing the performance in three tasks: handwritten digit recognition, language modelling and polyphonic music prediction. The results obtained from SNNs realized with the SNUs show competitive or better performance compared to the state-of-the-art ANNs. Note that in all these tasks we established the state-of-the-art performance for SNNs. Considering the sSNU, we demonstrate that for all three tasks a network with the simple sSNU dynamics surpasses the state-of-the-art performance of RNNs, LSTM- and GRU-based networks. Finally, we experimentally demonstrate highly-efficient in-memory acceleration of the synaptic operations of an SNU-based neural network using a phase-change memory (PCM) chip for music prediction. These results show that the SNU-based deep learning enables flexible and accurate training of SNNs in neuromorphic energy-efficient computing hardware.

## Spiking Neural Unit

Biological neural networks comprise neurons interconnected through synapses, which receive input spikes at the dendrites, and emit output spikes through the axons, as illustrated in Fig.1a. A neuronal membrane separates the intracellular space from the extracellular space. The membrane potential stays at a resting value in the absence of inputs and is altered by integrating the postsynaptic potentials through the dendrites. Upon sufficient excitation of the neuron, an action potential is generated as a result of currents transmitted through ion channels in the cell membrane[14]. Based on the Hodgkin and Huxley's abstraction of the neuronal membrane ion exchange dynamics to an RC electrical circuit[52], the spiking activity in SNN realizations is commonly modeled using the Leaky Integrate-and-Fire (LIF) neuron. It comprises a state variable $V_m$, corresponding to the membrane potential, with the dynamics described by the differential equation[14]

$$\frac{\mathrm{d}V_m(t)}{\tau} = -V_m(t) + RI(t), \tag{1}$$

or its discrete-time approximation assuming a discretization step $\Delta T$

$$V_{m,t+1} = \frac{\Delta T}{C} I_t + V_{m,t}\left(1 - \frac{\Delta T}{\tau}\right), \tag{2}$$

where $R$ and $C$ represent the resistance and the capacitance of the neuronal cell soma, respectively, $\tau = \mathrm{RC}$ is the time constant of the neuron, and $I(t)$ is the incoming current from the synapses. The synapses of a neuron $j$ receive spikes $x_i(t)$ and modulate them by the synaptic weights $W_{\mathrm{LIF},ji}$ to provide the input current to the neuronal cell soma. If we do not consider the temporal dynamics of biologically-realistic models of synapses and dendrites[14], the input current may be defined as $I_t = W_{\mathrm{LIF}} x_t$. The input current is integrated into the membrane potential $V_m$. When $V_m$ crosses a firing threshold $V_{th}$ at time $t$, an output spike is emitted: $y(t) = 1$, and the membrane potential is reset to the resting state $V_{rest}$, often defined to be equal to 0. This approach provides a basic framework for the analysis of the LIF dynamics, commonly explored in SNN research.

Here, we introduce a novel high-level model of the LIF dynamics, which we call a Spiking Neural Unit (SNU). The SNU comprises two ANN neurons as subunits: $N_1$, which models the membrane potential accumulation dynamics, and $N_2$, which implements the spike emission, as illustrated in Fig. 1b. The integration dynamics of the membrane-potential state variable is realized through a single self-looping connection to $N_1$ in the accumulation stage. The spike emission is realized through a neuron $N_2$ with step activation function. Simultaneously, an activation of $N_2$ controls the resetting of the state variable by gating the self-looping connection at $N_1$. Thus, SNU — a discrete-time abstraction of a LIF neuron — represents a construct that is directly implementable as a neural unit in ANN frameworks, where it may be scaled to deep architectures, as illustrated in Fig. 1c. Following the ANN convention, the formulas that govern the computations occurring in a layer of SNUs are as follows

$$\begin{aligned} s_t &= g\big(W x_t + l(\tau) \odot s_{t-1} \odot (1 - y_{t-1})\big) \\ y_t &= h(s_t + b), \end{aligned} \tag{3}$$

where $s_t$ is the vector of internal state variables calculated by the $N_1$ subunits, $y_t$ is the output vector calculated by the $N_2$ subunits, $g$ is the input activation function, $h$ is the output activation function, and $\odot$ denotes point-wise vector multiplication.

The activation function $g$ of $N_1$ may be implemented by the standard rectified linear activation function, if the common assumption that the membrane potential value is bounded by the resting state $V_{rest} = 0$ is adopted. However, alternative input activation functions can be used. The inputs are weighted by the synaptic weight matrix $W$ without a bias term. The self-looping weight $l(\tau)$, which is applied to the previous state value $s_{t-1}$, performs a discrete time approximation of the membrane potential decay that occurred in the time period $(t-1;t)$. If $l(\tau)$ is fixed to 1, the state does not decay and the SNU corresponds to the Integrate-and-Fire (IF) neuron without the leak term. The last term $(1-y_{t-1})$ relies on the binary output values of the spiking output to either retain the state, or reset it after spike emission. $N_2$ is a thresholding neuron, i.e. it has a step activation function $h(a)$, which returns 1 that corresponds to an output spike if $a > 0$, or 0 otherwise. There is no weight on the connection from $N_1$ to $N_2$ but it is biased with $b$ to implement the spiking threshold. These parameters correspond to the parameters of the LIF neuron introduced in Eq. 1 and the dynamics of SNU in an ANN framework corresponds to that of an LIF in an SNN framework, as summarized in Extended Data Fig. 1.

The ANN perspective on SNNs naturally sparks the idea of exploring alternative output activation functions. Traditionally, in SNNs information is transmitted throughout the network with all-or-none spikes, typically modeled as binary values. As a result, the input data is binarized, and the step activation function, $h$, of $N_2$ is used to determine the binary neuronal outputs. However, the proposed SNU implementation — within the ANN framework — allows the all-or-none constraint to be relaxed, thus allowing the benefits of a variant of the SNU, called soft

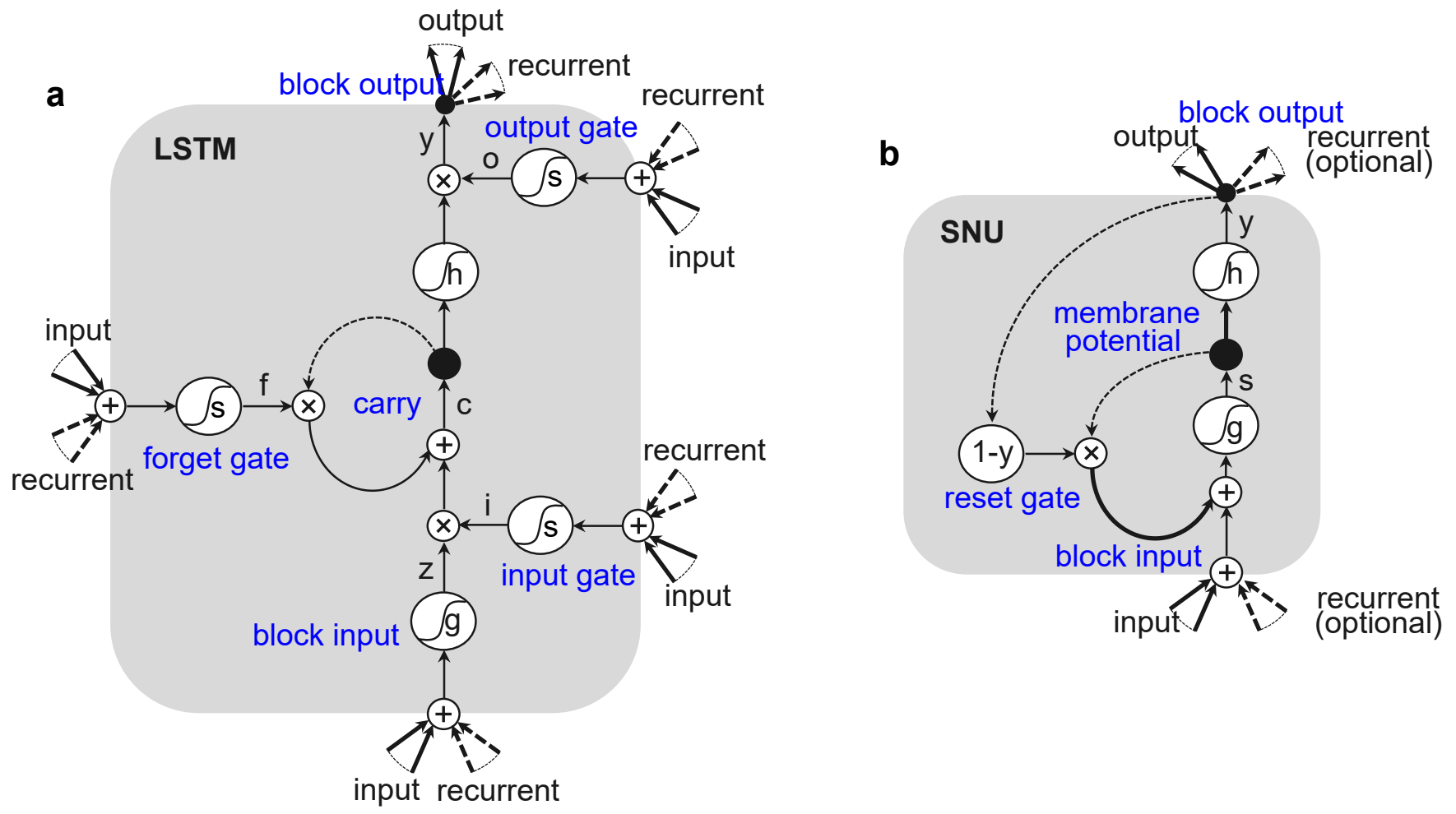

c

| | Network achitecture | |
|---|---|---|
| | Feed-forward | Recurrent |
| Stateless units | ANN | RNN |
| Stateful units | SNU-based | LSTM-/GRU-based |

d

| Neuron type | # of parameters |
|---|---|
| ANN neuron | $(m+1)n$ |
| RNN neuron | $(m+1)n+n^2$ |
| LSTM | $4((m+1)n+n^2)$ |
| GRU | $3((m+1)n+n^2)$ |
| SNU | $(m+1)n$ |
| recurrent SNU | $(m+1)n+n^2$ |

**Figure 2 | Analogies to the state-of-the-art ANNs. a**, LSTM drawing adapted from Greff et al.[62] Dashed lines indicate connections with time-lag. Bold lines indicate parametrized connections. **b**, SNU drawn based on the same convention. The ANN analogy enables to directly identify similarities and differences between other ANN units, such as LSTMs or GRUs. **c**, Structural classification of ANN architectures: SNUs espouse the common SNN approach of addressing temporal tasks using the feed-forward stateful architectures. **d**, Number of parameters for common ANN architectures in a single fully-connected layer with $n$ neurons and $m$ inputs.

SNU (sSNU) to be explored. The sSNU generalizes the dynamics to non-spiking ANNs, in which the input data does not have to be binarized and the activation function $h$ is set to a continuous function, such as a sigmoid. This formulation retains the temporal integration dynamics and has the additional interesting property of an analog proportional reset, i.e., the magnitude of the output determines what fraction of the membrane-potential state variable is retained. Exploiting the intermediate values at all stages of processing, viz., input, reset and output, facilitates on-par performance comparison of LIF-like dynamics with other ANN models, eliminating the impact of the limited value resolution of the standard SNU.

## The ANN-SNN duality

At the unit level, the temporal context of the SNU is maintained in its internal state, similarly to the state-of-the-art recurrent ANN units, such as LSTMs. In Figs. 2a-b, both units have a recurrent loop within the units' boundaries, drawn in gray. Their similarity becomes immediately apparent owing to fact that the SNU formulation enables the spiking neuron to be depicted using the ANN convention. Simultaneously, it enables the unique features of the SNU to be identified, viz., a non-linear transformation $g$ within the internal state loop, a parametrized state loop connection drawn in bold, a bias of the state output connection to the output activation function $h$ drawn in bold, and a direct reset gate $(1-y)$ controlled by the output $y$.

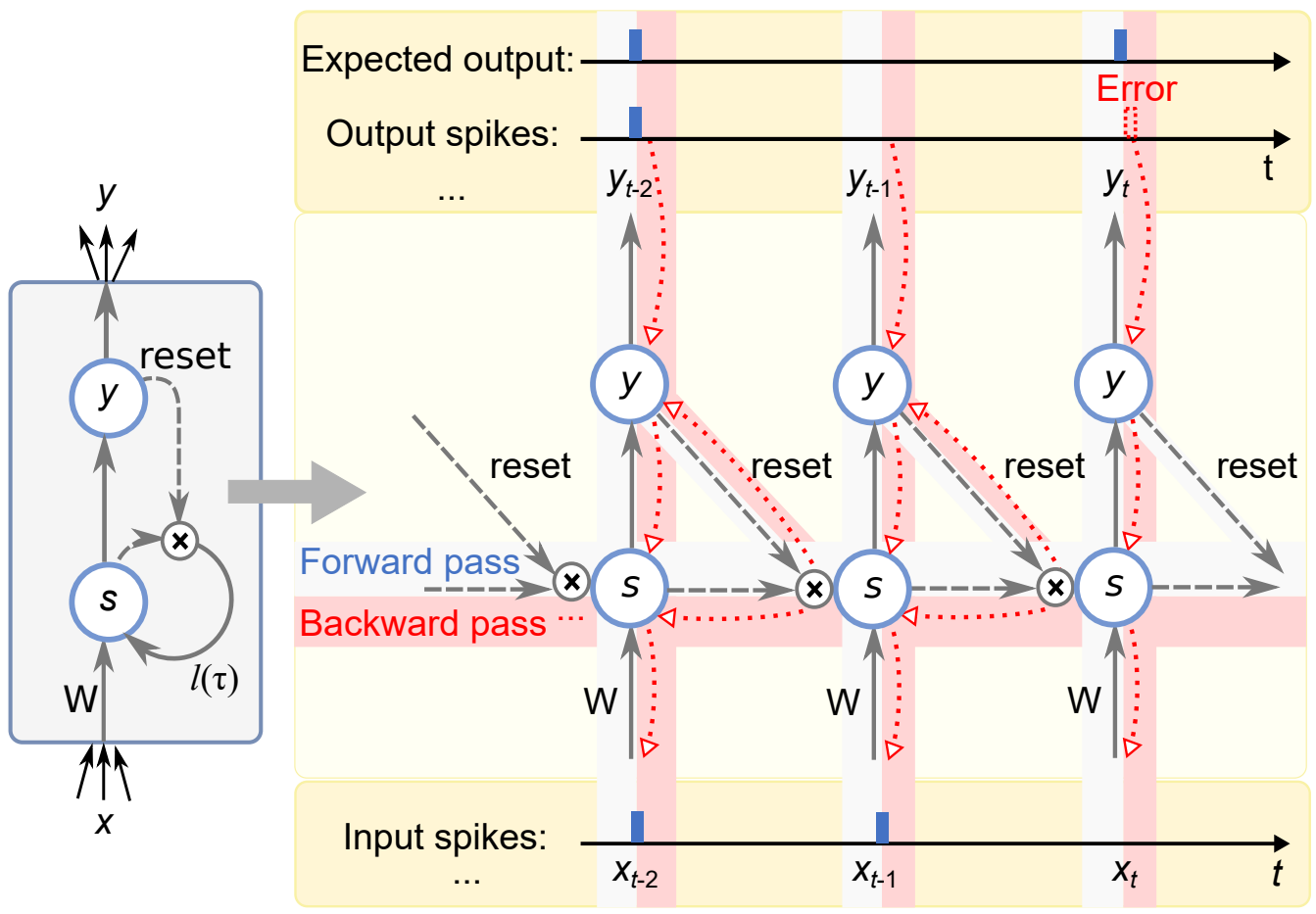

**Figure 3 | Training with backpropagation through time.** The temporal dynamics of SNU is unfolded over time during the forward pass, and error gradients are propagated backwards through the computational graph to determine the parameters' adjustments during the backward pass.

At the network level, SNUs can be optionally interconnected through recurrent connections, which might be beneficial for certain SNN architectures[18,50]. However, the standard approach in the SNN community is to rely on the internal state of feed-forward architectures[17,20,21,23,25–30,32]. The use of feed-forward stateful architectures for temporal problems has a series of profound advantages. From an implementation perspective, all-to-all connectivity between the neuronal outputs and the neuronal inputs within the same layer is not required. This may lead to highly-parallel software implementations or neuromorphic hardware designs. From a theoretical standpoint, a feed-forward network of SNUs is the simplest temporal architecture that creates a novel category of ANN architectures for temporal processing, as summarized in Fig. 2c. For a layer of $n$ neurons with $m$ inputs, the number of parameters for an SNU is $(m+1)n$ that is equivalent to the number of parameters in the simplest feed-forward ANN, summarized in Fig. 2d. SNU-based networks have the lowest number of parameters compared to the common temporal ANNs, viz., RNNs, GRU- or LSTM-based networks. Besides the reduced computational complexity, a low number of parameters results in faster training and may reduce overfitting.

Even though an SNU-based network is typically a feed-forward architecture, the state within the units is implemented using self-looping recurrent connections. For training such units, the ANN frameworks resort to algorithms such as backpropagation through time (BPTT)[53] . The mapping of the spiking neural dynamics to the standard ANN frameworks enables naturally the reuse of BPTT for training an SNU-based network. For this purpose, the SNU structure is unfolded over time, i.e., the computational graph and its parameters are replicated for each time step, as illustrated in Fig. 3, and then the standard backpropagation algorithm is applied. The unfolding involves only the local state of the neuron, which is different from the common RNN architectures that require unfolding of the activations of all units in a layer through recurrent connection matrices. Furthermore, for the computation of the gradients in the backward pass, backpropagation requires that all parts of the network are differentiable, which is the case for the sSNU variant, but not for the step function in the standard SNU. Nevertheless, in particular cases it is possible to train non-differentiable neural networks by providing a pseudo-derivative for the nondifferentiable functions[54]. Here, we follow this approach and use the derivative of tanh as the pseudo-derivative of the step function. In practice, the implementation of BPTT in well-established ANN frameworks generates a computational graph and uses automatic differentiation for the training, so that the entire training code is created dynamically. Therefore, owing to the SNU formulation, the SNNs can directly benefit from deep learning advancements, such as the training of deep multi-layer or convolutional architectures.

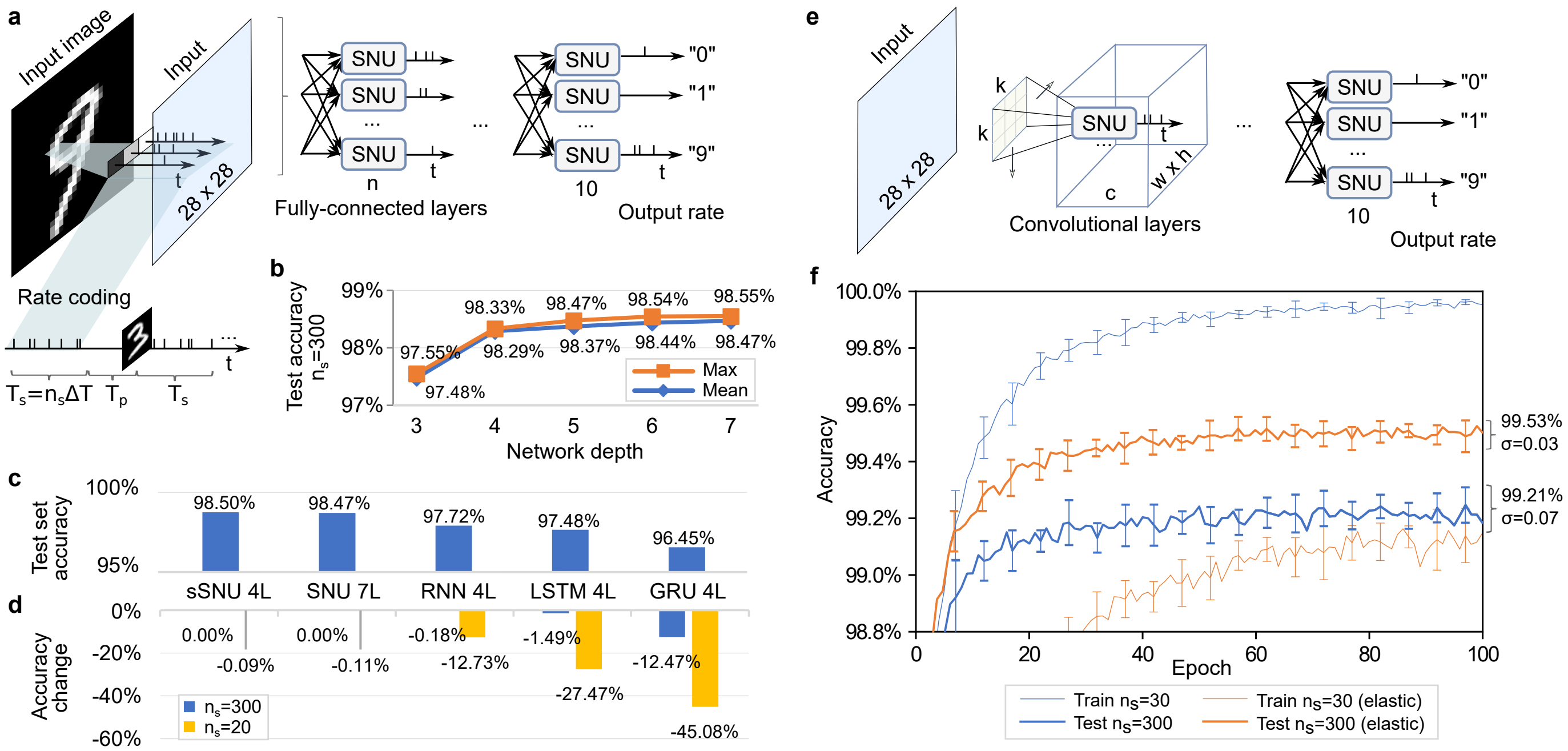

**Figure 4 | Application to rate-coded image classification.** The results in all panes of this figure are averaged over 10 different initial conditions. **a**, Grayscale input pixel value determines the rate of the input spikes, generated for a period $T_s$, corresponding to $n_s$ discrete-time steps of the SNU execution. A consecutive input image is commonly presented after the network activity decays during a pause $T_p$, but when stated we also consider a case with $T_p$=0. The inputs are processed by fully-connected layers, and the spiking activity at the last layer determines the assignment to one of ten classes. **b**, Fully-connected SNNs benefit from increased depth. **c**, SNU-based networks surpass the performance of common ANNs applied for the same task with the same architecture. 4-layer sSNU-based network obtains the best result. **d**, Accuracy change for testing with digits presented in the form of a continuous stream ($T_p$=0). SNU-based networks maintain high accuracy even for short presentation periods ($n_s$=20). **e**, Convolutional architecture with SNUs exploits the 2D characteristics of the images. **f**, Learning curves for SNU-based CNNs without and with elastic input distortions.

## Simulation results

We evaluated the performance of SNU-based networks on three tasks: handwritten digit recognition using MNIST dataset[55], language modelling using Penn Treebank (PTB) dataset[56] and polyphonic music prediction using Johan Sebastian Bach's chorales (JSB) dataset[57]. The inputs for the first task, which involves static data, were coded using the common SNN convention of rate-coding that enables the grayscale information of each pixel to be conveyed through a firing rate of the input spikes. The two other datasets were directly represented as sequences of binary inputs, so that they were directly fed into the spiking networks. Our goal was to compare the accuracies of SNU-based networks with the state-of-the-art SNNs as well as with common ANNs using similar training setups. For the SNU-based networks, the BPTT algorithm was applied for training.

For the handwritten digit recognition, we first evaluated the impact of the depth on fully-connected SNU-based SNNs, schematically illustrated in Fig. 4a. Following the insights from deep learning, we kept increasing the network depth, which has proven to be crucial for achieving higher accuracies. As illustrated in Fig. 4b, the mean recognition accuracy reached 98.47% for a 7-layer SNN. In comparison to various RNNs, LSTM- and GRU-based networks of similar architecture, the SNU-based networks achieved the highest accuracy, as illustrated in Fig. 4c. The best result of 98.5% was obtained using sSNUs in a 4-layer network. Furthermore, we assessed the generalization of the trained networks to a more challenging SNN evaluation convention[28,29]. Here, the inputs with consecutive digits formed a continuous stream, viz., the pause period was set to $T_p$=0 (see explanation at the bottom of Fig. 4a), and the network had to classify them without receiving explicit information when the digit at the input changed. Irrespective of the number of time steps $n_s$ for the test sequence, we observed almost no accuracy loss in SNU-based networks, contrary to up to over 45% loss for state-of-the-art ANNs receiving short sequences, as illustrated in Fig. 4d.

Secondly, to further improve the recognition accuracy and simultaneously to illustrate how the SNU concept directly benefits from the deep learning advancements, we implemented a convolutional SNU-based SNN, illustrated schematically in Fig. 4e, and with more details in Extended Data Fig. 2a. We used the same architecture, training setup and hyperparameters as in an ANN model[58], but applied it to the rate-coded MNIST dataset. After 100 epochs, we obtained an average accuracy of 99.21% for training with the original dataset and 99.53% for training with images preprocessed with elastic distortions[59], which act as a regularizer — decrease the training accuracy, but improve the test accuracy. The obtained accuracy beats that of the various state-of-the-art SNN implementations, as compared in Extended Data Figs. 2b-d.

The language modelling task involves predicting the next word based on the context of the previously observed sequence of words, schematically illustrated in Fig. 5a. Again, we reused insights from deep learning and built a common architecture with input embeddings and a softmax output layer[60]. A feed-forward SNU-based version of this architecture achieved perplexity of 137.7, which is better than traditional NLP approaches, such as 5-grams, as illustrated in Fig. 5b. To the best of our knowledge, this is the first ever example of language modelling performed with SNNs on the PTB dataset, and our result basically sets the SNN state-of-the-art performance. Application of sSNUs with recurrent connections improved the result down to 108.4, which surpassed the state-of-the-art ANNs without dropout, as illustrated in Extended Data Figs. 3a-b.

The task of polyphonic music prediction is to predict at each time step the set of notes that were to be played in the consecutive time step, schematically depicted in Fig. 5c. Similarly to the previous task, we used a hybrid architecture with an output layer of sigmoidal neurons, because it enabled the loss to be calculated in the same way as for ANNs and for the results to be compared with those of the state-of-the-art ANNs[61,62]. An SNU-based network achieved an average loss of 8.72 and set the SNN state-of-the-art performance for the JSB dataset. An sSNU-based network further reduced the average loss to 8.39 and surpassed state-of-the-art ANNs, as illustrated in Fig. 5d. The result was obtained with over 75% fewer parameters than the next result of a No Input Activation Function (NIAF) variant of LSTM, as illustrated in Extended Data Figs. 3c-d.

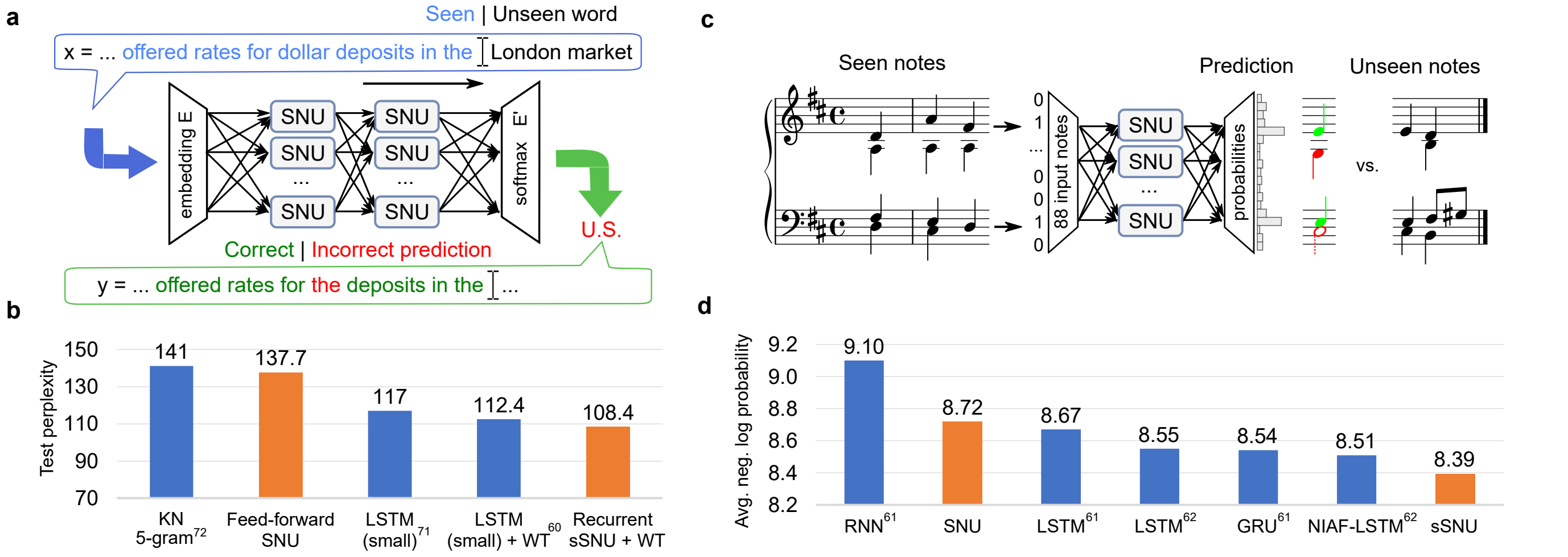

**Figure 5 | Application to sequence prediction.** **a**, Language modelling: the network predicts the consecutive words based on the context obtained from the past words. An actual example from SNU-based network is presented, where from the context of "dollar deposits" the network predicted that they should be in the "U.S.". Nevertheless, the ground truth is "London". **b**, Lower perplexity on the test set corresponds to higher-quality predictions of the model. **c**, Music prediction: the model predicts the probabilities of the consecutive notes that are to be played. **d**, The models are assessed using an averaged loss calculated over the predicted distributions.

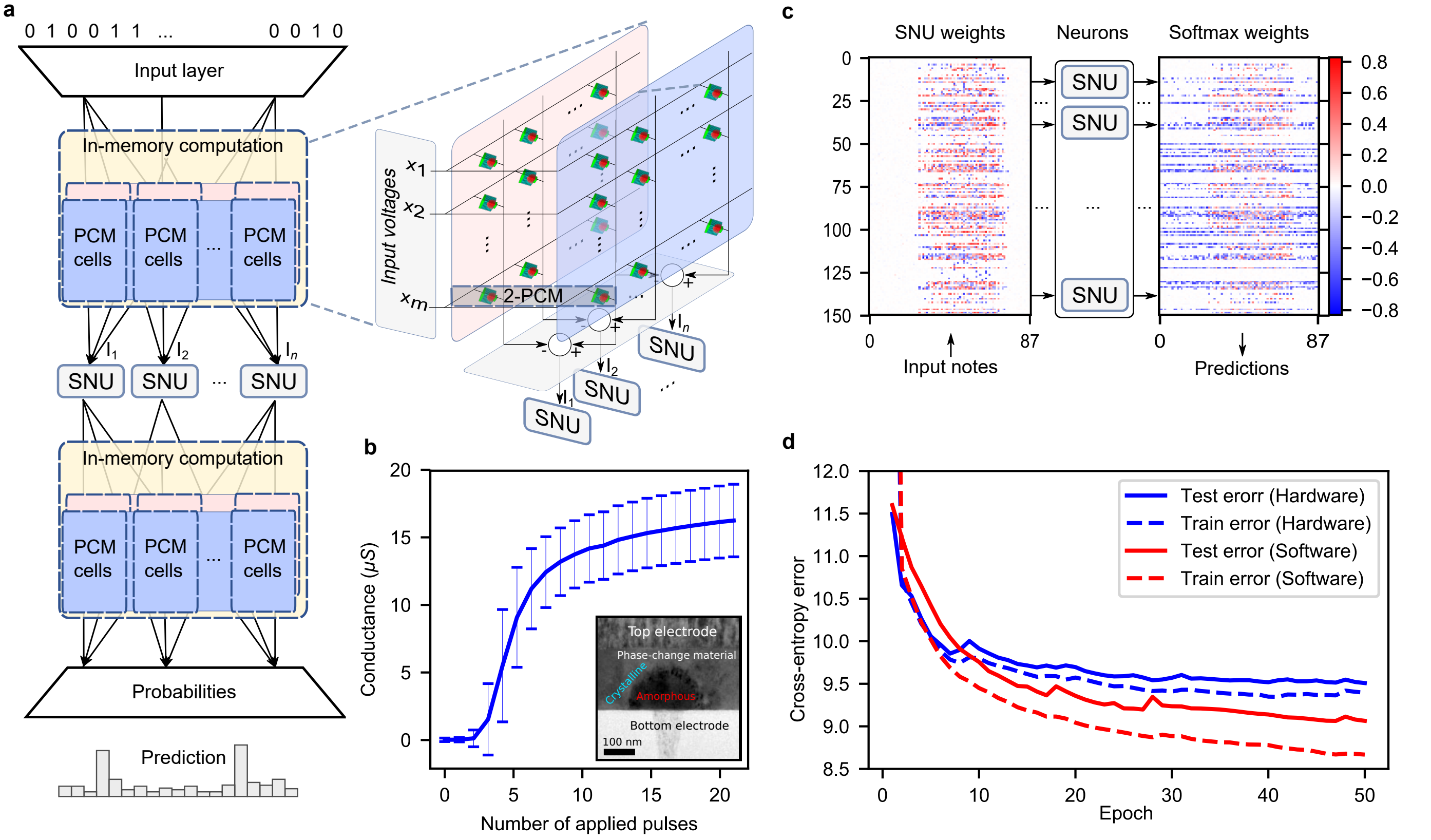

**Figure 6 | Neuromorphic in-memory acceleration.** **a**, The synaptic operations can be implemented with PCM crossbars that receive input spikes and calculate the synaptic input currents using physical properties of the architecture. We use two PCM devices per synapse in a 2-PCM differential configuration. **b**, The conductance of a PCM device depends on the relative configuration of the amorphous and crystalline phases. Accumulative characteristics of PCM device enables the gradual increase of the conductance with a series of crystallizing pulses. **c**, Final weights after the training. **d**, Learning curve for SNU-based music prediction network with hardware-in-the-loop and a software run with the same hyperparameters.

## Experimental demonstration in neuromorphic hardware

The use of ANN frameworks for SNNs implemented with SNUs can be further explored to enable operation on neuromorphic platforms. To illustrate this, we demonstrate the applicability of our approach to a hardware-in-the-loop architecture based on nanoscale phase-change-memory (PCM) devices, conceptually presented in Fig. 6a. The neuromorphic hardware concept consists of several crossbar memory arrays that map the synaptic weights to the conductances of the PCM cells. The PCM-based crossbar array structure enables highly area-efficient implementation of the synaptic weights and highly power-efficient in-memory calculation of the weighted inputs by avoiding to move the data from the memory to the computing units, which is required in contemporary von Neumann hardware architectures. Specifically, the memory cells perform the weighted input accumulation in-place based on their physical properties and the inherent capabilities of the crossbar structure. Crossbar-based acceleration of the multiply-accumulate enables its calculation with O(1) time complexity[63], potentially saving orders of magnitude in energy compared to current digital implementations of the neural network arithmetic operations. For reliable operation, in-the-loop training of the analog synaptic weights is essential to compensate the variations introduced by the reduced-precision in-memory computation.

We experimentally demonstrate the SNU-based hardware in-the-loop training by implementing the music prediction architecture from Fig. 5c in a prototype PCM-based platform[64] with an array of PCM devices. The synaptic inputs from the presynaptic neurons are converted to read voltage pulses sent to the PCM devices in the rows of the array. The resulting current at the columns of the array determines the input current $I$ to the postsynaptic neurons. For the remaining operations during the inference and training, the hardware-in-the-loop setup relies on the functionality of the ANN framework. During training, the framework communicates with the hardware to iteratively adjust the conductances of the PCM cells through application of programming pulses. The PCM cells are arranged in a 2-PCM differential configuration, marked in the right part of Fig. 6a, and illustrated in more detail in Extended Data Fig. 4a. The two PCM devices, $G+$ and $G-$, determine the effective synaptic weight in proportion to the difference of their conductances, $w_i \sim G_i+ - G_i-$. The $w_i$ weight increase is achieved by a $G_i+$ conductance increase whereas the $w_i$ weight decrease is achieved by a $G_i-$ conductance increase. This approach alleviates various practical issues stemming from the asymmetric programming characteristics of a PCM cell, illustrated in Fig. 6b. The conductance of the cell increases by gradual shrinking of the amorphous dome through application of low-power crystallizing pulses, which enables gradual weight updates $\Delta w$ to be implemented during the training. The conductance decrease requires the amorphous phase to be recreated through the melting process with higher energy pulses, typically referred to as reset pulses. In the 2-PCM cell approach reset pulses are only applied during cyclic conductance rebalancing to avoid saturation of the $G_i+$ and $G_i-$ conductances[65].

To train the SNU-based network for the music prediction task, we mapped the 26400 synaptic weights to 52800 PCM devices. The network was trained with BPTT on the JSB dataset for 50 epochs. During training, the hardware weights were read and their values were updated by adjusting the device conductances through programming pulses. The resulting weights of the two trainable layers of the network after learning are demonstrated in Fig. 6c, and the evolution of their weight distributions is illustrated in Extended Data Fig. 4b. The learning curves are depicted in Fig. 6d. The final result we obtained in hardware is 9.51 and in software using the same hyperparameters is 9.08.

This experimental evaluation demonstrates the robustness of the SNU-based training when the network comprises noisy and stochastic on-chip synapses. Furthermore, these results provide the pathway to extend the applicability of memristive-based neuromorphic hardware to supervised learning from temporal datasets. For instance, after training with SNUs, the hardware could be disconnected from the software, and the synaptic weights stored in the chip could be used for highly-efficient all-in-memory inference.

## Conclusion

For a long time, SNN and ANN research and applications have been developing separately. There has been significant effort to harness the unique capabilities of the SNN dynamics, albeit with limited success compared to the spectacular progress witnessed in deep learning of ANNs. In this paper, we bridged these neural network architectures by proposing an SNU that incorporates the biologically-inspired spiking neural dynamics in form of a novel ANN unit, and further generalizes it to a non-spiking case in the soft SNU.

The transformation of the spiking model to the SNU leads to ANN-SNN duality that enables to directly compare the dynamics of the spiking neural units with the state-of-the-art ANNs. This dynamics provides different capabilities than that of RNNs, LSTM- or GRU-based networks, and requires the fewest parameters per neuron among the existing ANN models for temporal data processing. Secondly, it enables to reuse standard ANN frameworks for SNN implementation and training with BPTT without the need to derive SNN-specific solutions. Thirdly, it enables to directly benefit from the deep learning advances and seamlessly integrate SNUs into deep learning architectures.

The SNU advantages were demonstrated in a series of benchmarks on three tasks, in all of which our approach set the state-of-the-art performance for SNNs, and in the soft variant, surpassed ANN performance of similar networks. For rate-coded handwritten digit recognition, SNUs with up to 7 layers outperformed state-of-the-art ANNs and demonstrated better robustness to a more challenging continuous input presentation scheme. Direct incorporation of SNUs into a convolutional ANN architecture further improved the results. For language modelling, we again reused ANN architectures and demonstrated that SNU-based networks perform better than a classic NLP model. For music prediction, we demonstrated competitive performance of SNNs and high performance of sSNUs that surpassed the results of the state-of-the-art LSTMs. Finally, we experimentally demonstrated the applicability of SNU-based approach to neuromorphic hardware. We performed training of PCM-based arrays of memristive synapses in a prototype chip. The obtained synaptic configurations achieved competitive results and remained in the chip, that potentially can be used as a core of a highly-efficient in-memory non-von-Neumann inference machine.

The proposed SNU opens many new avenues for future work. It enables to explore the capabilities of biologically-inspired neural models and benefit from their low computational power as well as their simplicity. It also provides an easy approach to training spiking networks that could increase their adoption for practical applications and would unlock the benefits of power-efficient neuromorphic hardware implementations. Finally, the compatibility of the SNU with the ANN frameworks and models enables the use of existing or forthcoming ANN accelerators for SNN implementation and deployment.

**a**

| LIF nomenclature | Correspondence | SNU nomenclature |
|---|---|---|
| membrane potential | $V_m = s$ | unit state |
| spiking threshold | $V_{th} = -b$ | bias |
| input soma current | $I$ | |
| | $\Delta T/C\ I = Wx$ | weighted input |
| membrane time constant | $\tau$ | |
| | $(1-\Delta T/\tau) = l(\tau)$ | state decay |

**b**

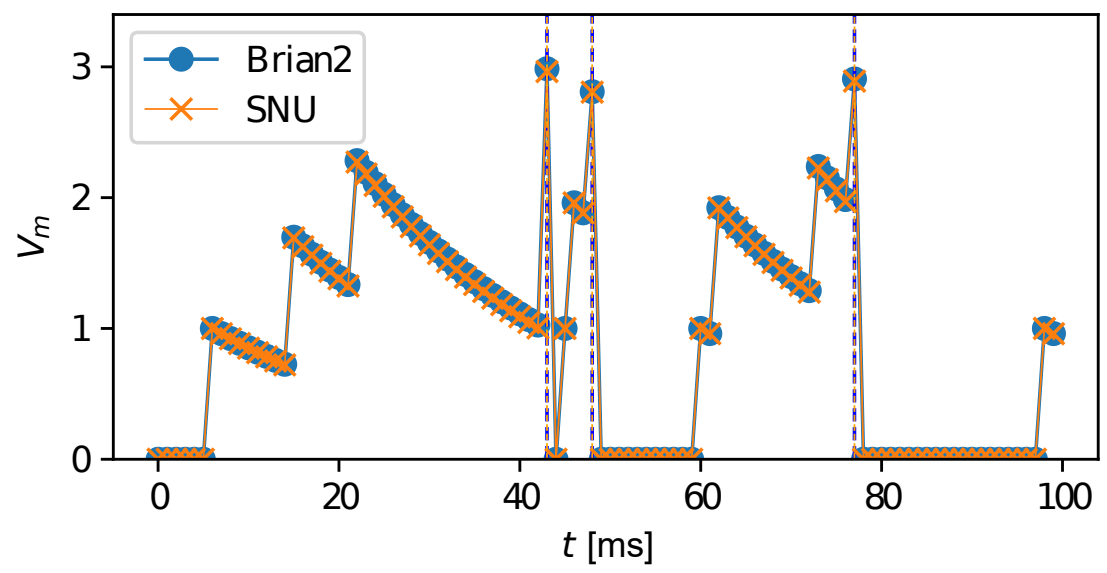

**Extended Data Figure 1 | Correspondence between an SNU and an LIF neuron. a**, The respective LIF parameters directly correspond to the SNU parameters, such that the same set of parameter values can be used in an SNU-based network, implemented by utilizing standard ANN frameworks, as well in a native LIF-based implementation, utilizing standard SNN frameworks. **b**, To demonstrate this, we have used TensorFlow[66] to produce sample plots of the spiking dynamics for a single SNU. The state variable of the SNU increases each time an input spike arrives at the neuron, and decreases following the exponential decay dynamics. When the spiking threshold is reached, an output spike is emitted (vertical dashed line) and the membrane potential is reset. These dynamics are aligned with the reference LIF dynamics, which we obtained for the corresponding parameters by running a simulation in the well-known Brian2[67] SNN framework.

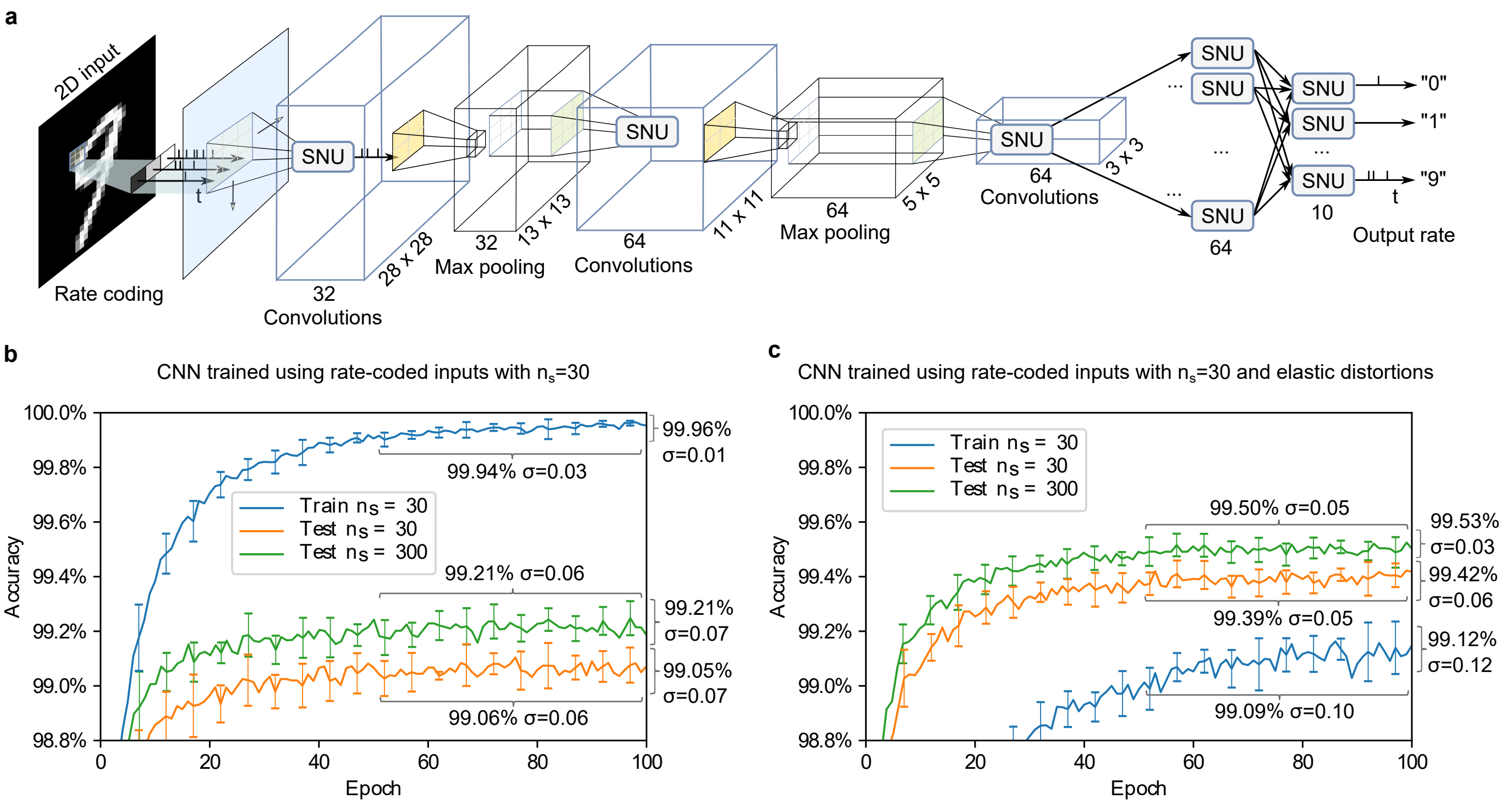

**d**

| Type | Comments | Parameters | Accuracy | Reference |
|---|---|---|---|---|
| FC 2L | STDP | 5,017,600 | 95.00% | Diehl & Cook, 2015 |
| CNN | SNN gradient descent; fixed Gabor kernels | 81,120 | 98.17% | Kulkarni & Rajendran, 2018 |
| FC 4L | ANN to SNN conversion | 494,710 | 98.37% | Hunsberger & Eliasmith, 2015 |
| CNN | STDP | 5,875,456 | 98.40% | Kheradpisheh et al., 2018 |
| FC 7L | SNUs using BPTT with SGD | 466,698 | **98.47%** | this work |
| FC 3L | SNN using BP with ADAM; lateral inhibition; regularization | 328,984 | 98.77% | Lee J.H. et al., 2016 |
| CNN | SNUs using BPTT with RMSProp | 93,322 | **99.21%** | this work |
| CNN | STDP pre-training; BP | 277,780 | 99.28% | Lee C. et al., 2018 |
| CNN | SNN using BP with ADAM; regularization; elastic distortions | 581,520 | 99.31% | Lee J.H. et al., 2016 |
| CNN | SNN using BSTP; elastic distortions | 318,740 | 99.42% | Wu et al., 2018 |
| CNN | SNUs using BPTT with RMSProp; elastic distortions | 93,322 | **99.53%** | this work |

**Extended Data Figure 2 | Image classification details. a**, Complete spiking CNN architecture. **b**, CNN learning curve for rate-coded inputs without preprocessing. The accuracy was calculated by averaging over 10 different initializations (vertical brackets) or also over the last 50 epochs (horizontal brackets). **c**, Analogous CNN learning curve for rate-coded inputs obtained from MNIST images preprocessed with elastic distortions. **d**, Table comparing the state-of-the-art fully-connected (FC) and convolutional (CNN) SNN architectures[27,42,48,51,68–70] in terms of parameters and obtained MNIST accuracy.

a

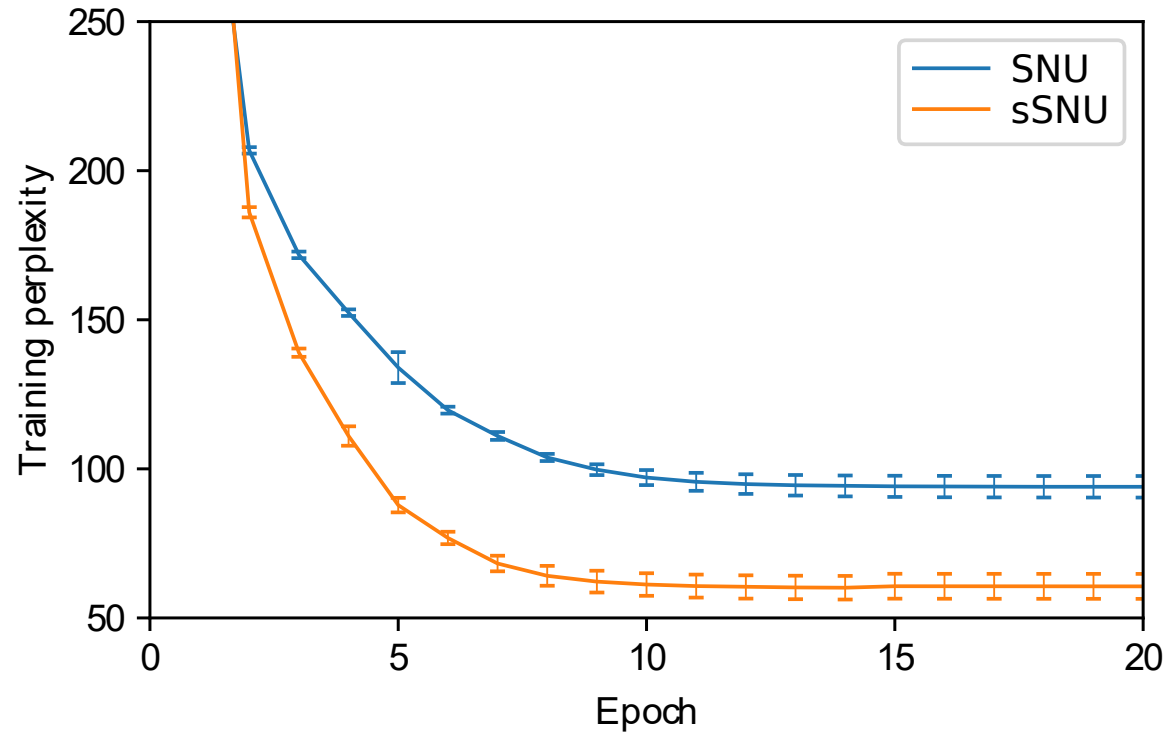

c

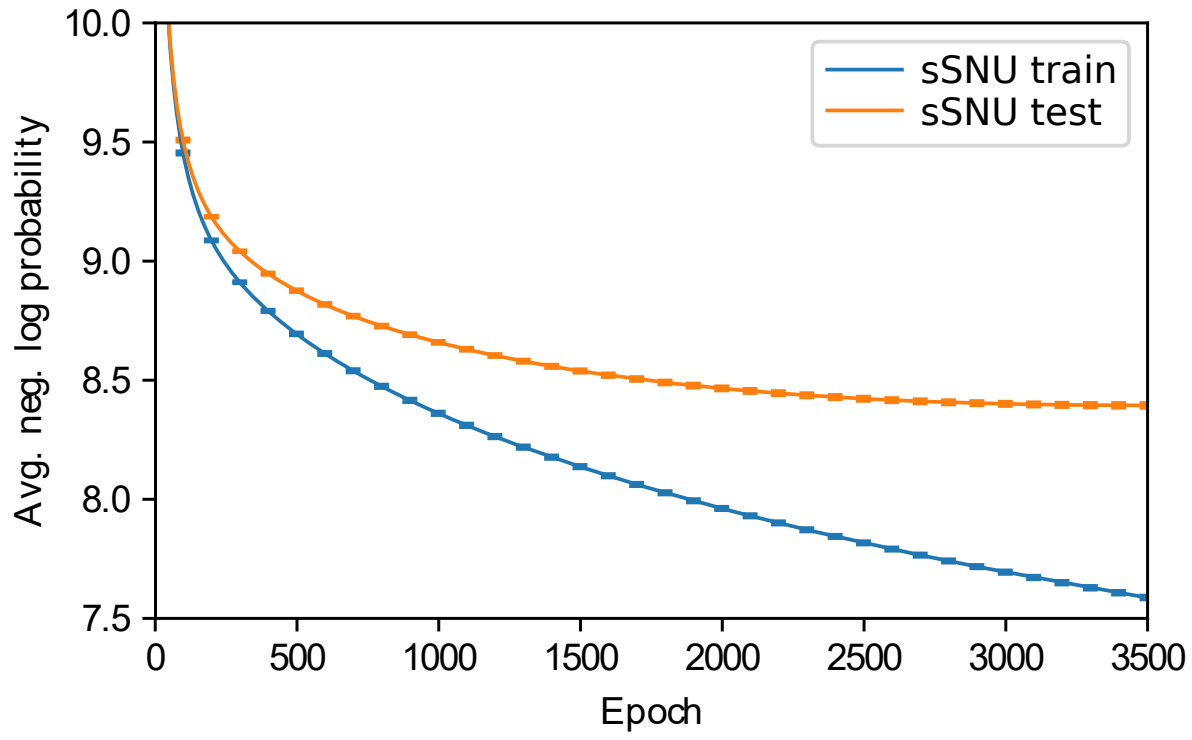

b

| Model | Parameters | Perplexity | Reference |
|---|---|---|---|
| KN 5-gram | | 141 | Mikolov et al., 2011 |
| Feed-forward SNU | 4.65M | **137.7 σ=1.6** | this work |
| RNN | | 123 | Mikolov et al., 2011 |
| LSTM (small) | 4.65M | 117 | Zaremba et al., 2014 |
| LSTM (small) + WT | 2.65M | 112.4 | Press & Wolf, 2017 |
| Recurrent sSNU + WT | 19.8M | **108.4 σ=1.0** | this work |

d

| Model | Parameters | Avg. neg. log prob. | Reference |
|---|---|---|---|
| RNN | 30.2k | 9.10 | Chung et al., 2014 |
| SNU | 26.6k | **8.72 σ=0.04** | this work |
| LSTM | 23.5k | 8.67 | Chung et al., 2014 |
| LSTM | 69k | 8.55 (best 10% runs) | Greff et al., 2017 |
| GRU | 25k | 8.54 | Chung et al., 2014 |
| NIAF-LSTM | 97.3k | 8.51 (best 10% runs) | Greff et al., 2017 |
| sSNU | 26.6k | **8.39 σ=0.01** | this work |

**Extended Data Figure 3 | Sequence prediction details.** The values in all the panes of this figure were obtained by averaging over 10 different initializations. Standard deviation is reported along the results and marked with error bars in the plots. **a**, Language modelling training perplexity evolution for SNU- and sSNU-based architectures. **b**, Comparison of test perplexity with other results[60,71,72]. ANN results using standard architectures with similar training techniques were considered, i.e. no pre- or post-processing, single network, truncated BPTT, no dropout. WT denotes weight tying of the output layer with the embedding layer. **c**, Music prediction loss evolution for sSNU-based network. **d**, Comparison with other results[61,62].

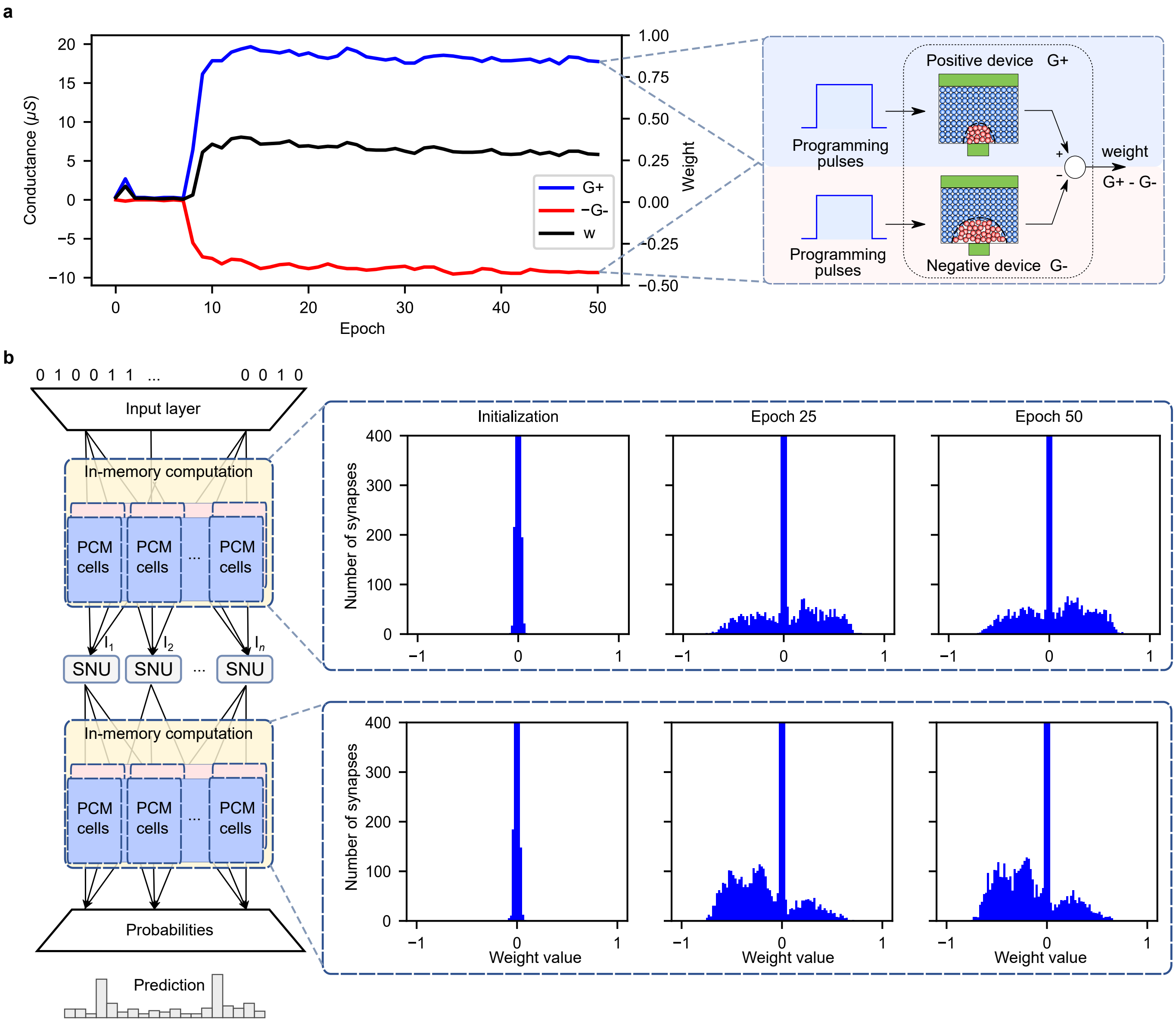

**Extended Data Figure 4 | Hardware experiment details.** **a**, A 2-PCM synapse is implemented with 2 PCM devices operating in a differential configuration, i.e. a weight w is proportional to a difference between the conductances of the G+ and the G- device. Weight increase is performed through crystallization of the positive device with programming pulses and weight decrease is performed through crystallization of the negative device with programming pulses. The plot on the left contains an example evolution of the 2-PCM synapse over the course of training. Aside from programming pulses, the fluctuations in the conductance values arise owing to PCM-specific physical phenomena, such as read noise or conductance drift. **b**, Snapshots of the weight distributions over the course of training, depicted for the two trainable layers.